\documentclass[
]{ceurart}

\usepackage{listings}
\usepackage{graphicx}
\usepackage{subcaption}
\usepackage{placeins}  

\begin{document}

\copyrightyear{2025}
\copyrightclause{Copyright for this paper by its authors.
  Use permitted under Creative Commons License Attribution 4.0
  International (CC BY 4.0).}

\conference{AIP2025: Second Workshop on AI in Production, 
  2025}

\title{Teacher-Student Guided Inverse Modeling for Steel Final Hardness Estimation}

\author[1,2]{Ahmad Alsheikh}[%
email=ahmad.alsheikh@krones.com,
]
\author[2]{Andreas Fischer}[%
email=andreas.fischer@th-deg.de,
]

\address[1]{KRONES AG, Böhmerwaldstr. 5, 93073 Neutraubling, Germany}
\address[2]{Deggendorf Institute of Technology, Dieter-Görlitz-Platz 1, 94469 Deggendorf, Germany}

\begin{abstract}
Predicting the final hardness of steel after heat treatment is a challenging regression task due to the many-to-one nature of the process—different combinations of input parameters (such as temperature, duration, and chemical composition) can result in the same hardness value. This ambiguity makes the inverse problem—estimating input parameters from a desired hardness—particularly difficult. In this work, we propose a novel solution using a Teacher-Student learning framework. First, a forward model (Teacher) is trained to predict final hardness from 13 metallurgical input features. Then, a backward model (Student) is trained to infer plausible input configurations from a target hardness value. The Student is optimized by leveraging feedback from the Teacher in an iterative, supervised loop. We evaluate our method on a publicly available tempered steel dataset and compare it against baseline regression and reinforcement learning models. Results show that our Teacher-Student framework not only achieves higher inverse prediction accuracy but also requires significantly less computational time, demonstrating its effectiveness and efficiency for inverse process modeling in materials science.
\end{abstract}


\maketitle

\section{Introduction}
Heat treatment is a cornerstone in materials engineering for modifying the mechanical properties of metals, with hardness being a critical target metric in structural and industrial applications. Processes such as tempering, quenching, and annealing involve multiple input parameters including temperature, treatment time, and chemical composition which interact in nonlinear and often material-specific ways. The resulting complexity makes it difficult to predict the process conditions required to achieve a desired final hardness.

A particularly challenging aspect of this problem is its many-to-one nature: Multiple combinations of input variables can yield the same hardness result. This poses a fundamental obstacle to the inverse prediction, where the goal is to infer the set of input conditions that will produce a specified output. Mathematically, many-to-one mappings are non-invertible. This ambiguity complicates the application of traditional regression techniques in inverse modeling. Addressing this challenge requires modeling strategies that can incorporate solution multiplicity, enforce output consistency, and remain robust to the inherent uncertainty of inverse mapping.

This ambiguity complicates the application of traditional regression techniques in inverse modeling, as such methods typically assume a one-to-one correspondence between inputs and outputs. Moreover, standard optimization methods often struggle to converge reliably when faced with non-invertible mappings, as the error surface can contain multiple local minima or flat regions corresponding to different valid inputs. Addressing this challenge requires modeling strategies that can incorporate solution multiplicity, enforce output consistency, and remain robust to the inherent uncertainty of inverse mapping.

The purpose of this study is to explore and address the inverse problem of predicting process parameters for steel heat treatment under the constraints of many-to-one mappings. By examining this problem in depth, we aim to better understand the limitations of existing approaches and identify avenues for constructing more ro-bust, generalizable inverse models within materials informatics.

The remainder of this paper is organized as follows. Section 2 reviews prior research related to non-invertible mappings and inverse modeling techniques in machine learning. Section 3 presents our proposed framework. Section 4 outlines the experimental setup and results. Section 5 concludes with key findings and potential future directions.

\section{Related Work}

In this section, we review recent advancements addressing the challenges of One-to-Many and Many-to-One mappings in machine learning, specifically in regression tasks. These mappings often introduce ambiguity due to their non-invertible nature, which complicates prediction, generalization, and interpretation. To address these challenges, researchers have explored a variety of modeling techniques, optimization strategies, and auxiliary frameworks.

Grollman and Jenkins \cite{grollman2009} proposed a multi-map regression approach to handle perceptual aliasing in robotic controllers. Their method employs sparse online learning to resolve ambiguities introduced by overlapping many-to-one mappings, demonstrating its effectiveness in robotics applications. Courts and Kvinge \cite{courts2021} introduced bundle networks, a framework that uses generative modeling and fiber bundles to disentangle ambiguities in classification and regression tasks by generating local trivializations.

In robotics, Singh et al. \cite{singh2022} used regression-based kinematic modeling to optimize gait trajectories for biped robots, incorporating auxiliary constraints to improve the handling of many-to-one mappings. Valdés and Tchagang \cite{valdes2020} tackled inverse mappings in simulation-based models by combining deterministic surrogate models with machine learning approaches, highlighting the strength of mixed strategies in regression.

Yang et al. \cite{yang2017} used recurrent neural networks to model therapy decision making in metastatic breast cancer, addressing ambiguity through hierarchical regression and encoder-decoder architectures. Chen and Zhu \cite{chen2019} proposed a guided deep learning algorithm for structural surface design, showing that incorporating multiple loss functions can help disambiguate many-to-one regression outputs. Kreuzig et al. \cite{kreuzig2019} developed DistanceNet, which combines recurrent convolutional neural networks with ordinal regression to estimate traveled distance from monocular images, effectively reducing mapping ambiguity in visual tasks.

In the medical imaging domain, Yurt et al. \cite{yurt2021} introduced mustGAN, a multi-stream generative adversarial network for MR image synthesis. Their model addresses feature-level ambiguities through adversarial training. Wang et al. \cite{wang2024} presented M2ORT, a transformer-based framework for spatial transcriptomics prediction, demonstrating how auxiliary data and novel architectural designs can improve the modeling of non-invertible mappings.

Zhang et al. \cite{zhang2022} proposed a local-to-global cost aggregation method for semantic correspondence using a Teacher-Student framework. This approach adapts learned representations to mitigate ambiguities in many-to-one feature matching tasks. Lastly, Neupane et al. \cite{neupane2024} surveyed techniques for 3D human pose estimation, focusing on methods that apply auxiliary constraints and deep learning models to resolve depth ambiguities inherent in coordinate regression.

\section{Methodology: Teacher-Student for Inverse Prediction}

Solving inverse problems in many-to-one mappings is especially challenging due to the inherent ambiguity: a single output can correspond to multiple valid input configurations. Traditional regression methods struggle with this non-invertibility, often resulting in poor generalization and conflicting gradients during optimization. To overcome these limitations, we propose a Teacher-Student learning framework for robust inverse prediction.

\begin{figure}[t!]
  \centering
  \begin{subfigure}[t]{0.4\linewidth}
    \centering
    \includegraphics[width=\linewidth]{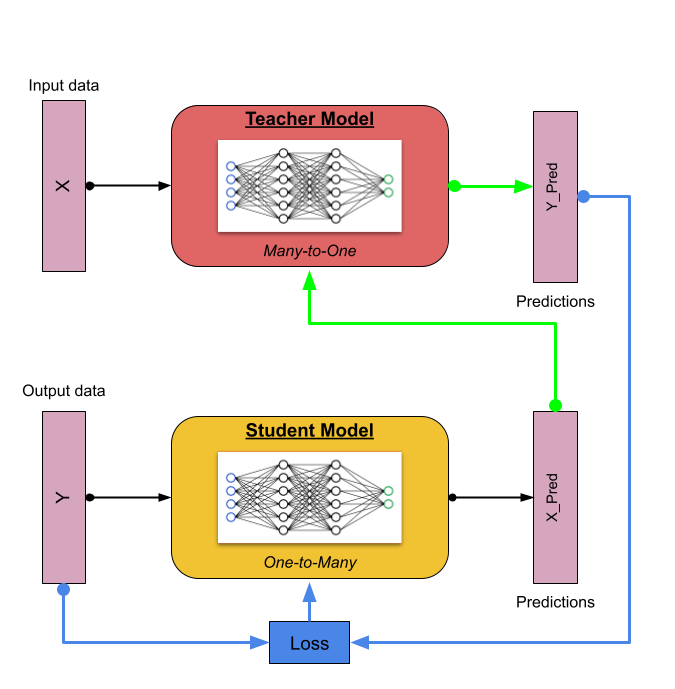}
    \caption*{(a)}
  \end{subfigure}
  \hfill
  \begin{subfigure}[t]{0.59\linewidth}
    \centering
    \includegraphics[width=\linewidth]{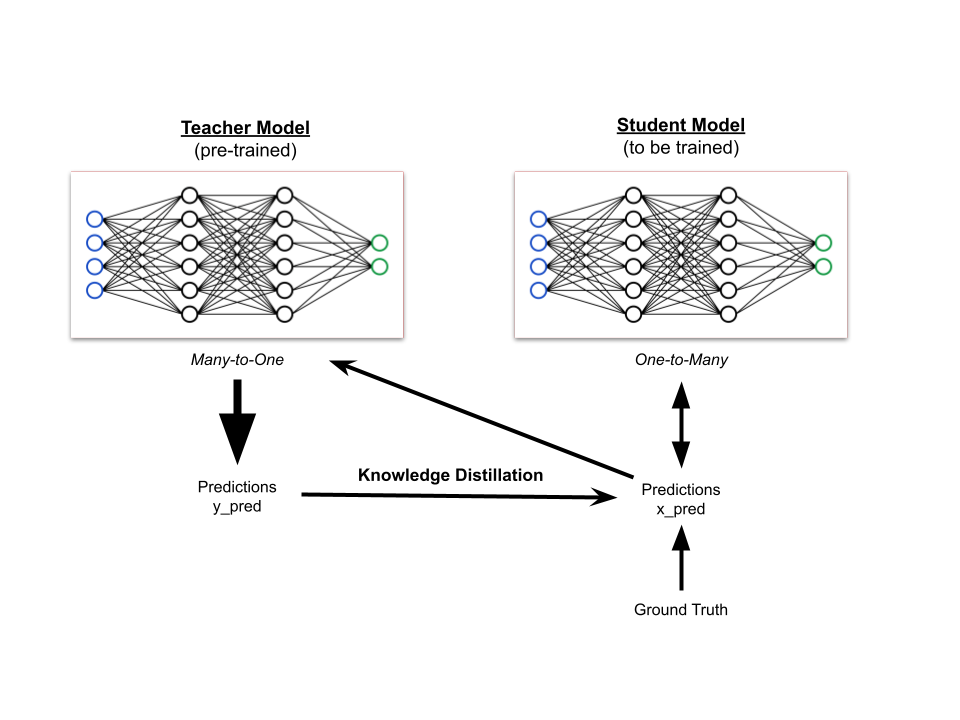}
    \caption*{(b)}
  \end{subfigure}
  \caption{Overview of the Teacher-Student framework for inverse hardness prediction. (a) The Teacher model is trained to map process inputs ($X$) to final hardness ($Y$). The Student model learns to predict valid input configurations ($X_{\text{pred}}$) from target hardness values. Predictions are evaluated by the fixed Teacher model, and the loss is used to iteratively update the Student. (b) A schematic view highlighting the knowledge distillation process between Teacher and Student models.}

  \label{fig:architecture}
\end{figure}

As illustrated in Figure~\ref{fig:architecture}, the framework consists of two Multi-Layer Perceptron (MLP) models:

\begin{itemize}
  \item A \textbf{Teacher model}, trained on the forward (many-to-one) task, maps a set of 13 input features—including tempering time, temperature, and elemental composition—to the final hardness value (HRC).
  \item A \textbf{Student model}, trained on the inverse (one-to-many) task, predicts plausible input configurations given a target hardness.
  \item Both models use three dense layers with ELU activations and residual connections.
\end{itemize}

The training process proceeds in two phases. First, the Teacher model is trained to accurately learn the forward mapping using supervised learning. Once trained, it remains fixed and acts as a reference model for supervising the Student.

The Student model receives randomly sampled target hardness values as inputs and predicts the corresponding input parameters. These predicted inputs are then passed through the Teacher model, which outputs a predicted hardness. This predicted value is compared to the original target, and a loss is computed. The loss is then back-propagated to update the Student model's weights. This process continues iteratively, allowing the Student to learn input patterns that are functionally consistent with the desired hardness values.

Importantly, because of the many-to-one nature of the problem, the Student is not trained to replicate original dataset entries. Instead, it learns to generate valid, functionally equivalent inputs—those that the Teacher model accepts as leading to the correct hardness. This design enables the system to embrace ambiguity while still producing accurate and interpretable predictions for inverse process modeling.

\section{Experimental Setup and Results}

\subsection{Dataset Overview}

To evaluate our proposed framework, we used a publicly available dataset \cite{raiipa2024} containing steel samples subjected to various heat treatment conditions. Each data point consists of 13 input features:

\begin{itemize}
  \item \textbf{Process parameters:} tempering time and temperature
  \item \textbf{Chemical composition:} elemental percentages of C, Mn, P, S, Si, Ni, Cr, Mo, V, Al, and Cu
\end{itemize}

The target variable is the final hardness of the steel sample, measured in Rockwell Hardness (HRC) after quenching and tempering.

Figure~\ref{fig:many-to-one}a illustrates the relationship between tempering temperature and final hardness (HRC) for varying tempering times. As temperature increases, final hardness consistently decreases across all time intervals, confirming a strong inverse correlation. Longer tempering durations (e.g., 100{,}000 s) result in lower hardness compared to shorter durations (e.g., 20{,}000 s), highlighting the combined effect of thermal exposure and time on material softening. Elements like Mn and P show positive correlation, while others (e.g., C, Cr, Si) exhibit weaker or nonlinear relationships.

\subsection{Observing the Many-to-One Mapping}

\begin{figure}
  \centering
  \begin{subfigure}[t]{0.51\linewidth}
    \centering
    \includegraphics[width=\linewidth]{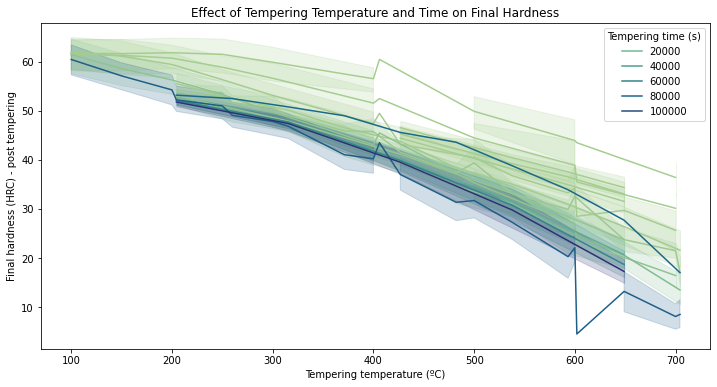}  
    \caption*{(a)}
  \end{subfigure}
  \hfill
  \begin{subfigure}[t]{0.44\linewidth}
    \centering
    \includegraphics[width=\linewidth]{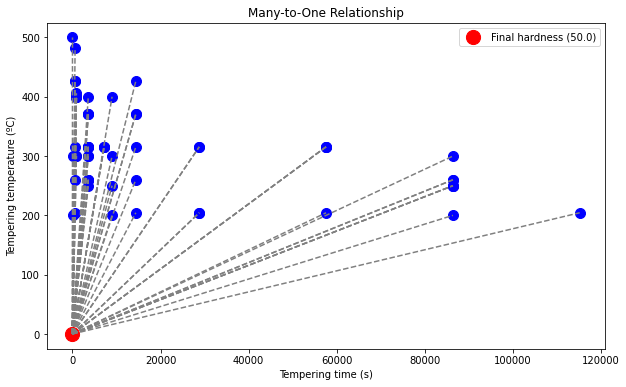}  
    \caption*{(b)}

  \end{subfigure}
  \caption{Exploratory data analysis. (a) Relationship between tempering temperature, time, and final hardness (HRC). Higher temperatures and longer durations result in lower hardness, illustrating a clear inverse trend. (b) Multiple input configurations leading to the same hardness value, highlighting the many-to-one nature of the inverse problem.}

  \label{fig:many-to-one}
\end{figure}

Exploratory analysis revealed that the dataset exhibits a many-to-one mapping: multiple distinct combinations of input parameters lead to the same hardness value. This is visualized in Figure~\ref{fig:many-to-one}b, where overlapping input regions map to identical target values. Such non-invertibility complicates inverse prediction, as standard regression models struggle to differentiate between equally valid input solutions.

\subsection{Baseline Inverse Modeling Attempts}

To assess the complexity of the inverse prediction task, we first trained two conventional regression models to estimate process parameters from a given target hardness value:

\begin{itemize}
  \item A Random Forest Regressor
  \item A Multi-Layer Perceptron (MLP)
\end{itemize}

Both models were configured to take the final hardness (HRC) as input and predict the full set of 13 output variables, including tempering parameters and elemental composition. Hyperparameters were optimized using random search.

\begin{figure}[t]
  \centering
  \begin{subfigure}[t]{0.44\linewidth}
    \centering
    \includegraphics[width=\linewidth]{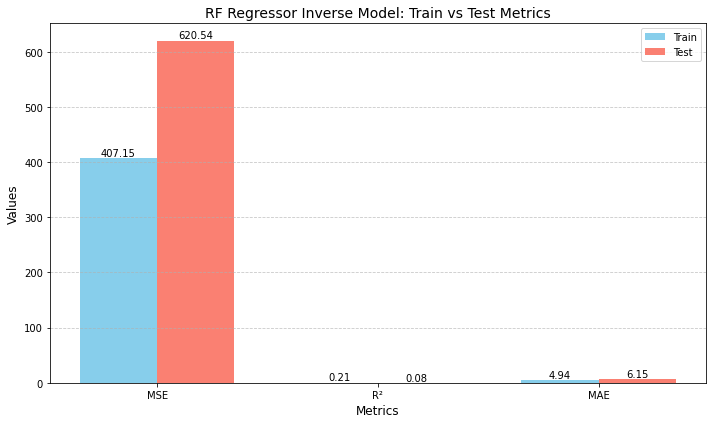}
    \caption*{(a)}
  \end{subfigure}
  \hfill
  \begin{subfigure}[t]{0.52\linewidth}
    \centering
    \includegraphics[width=\linewidth]{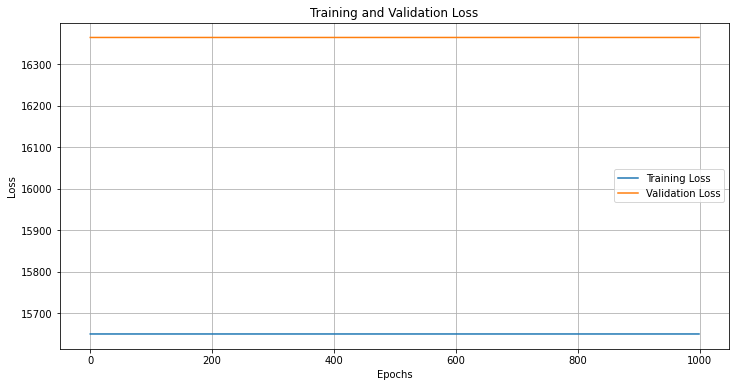}
    \caption*{(b)}
  \end{subfigure}
  \caption{Baseline model performance. (a) The Random Forest inverse model shows high test error and low generalization. (b) The MLP baseline failed to converge, with persistent high loss across training and validation. These results highlight the difficulty of solving many-to-one inverse prediction problems with standard models.}
  \label{fig:baseline}
\end{figure}

Despite optimization, the results demonstrate the limitations of using off-the-shelf regression models for one-to-many problems. The random forest regressor yielded high prediction error, with an MSE of 620.54 on the test set and a low $R^2$ score of 0.08, as shown in Figure~\ref{fig:baseline}a. This large discrepancy between predicted and true values highlights the model’s inability to generalize in the presence of multiple valid input solutions for the same output.

The MLP model also performed poorly. As seen in Figure~\ref{fig:baseline}b, both training and validation loss remained flat across 1000 epochs, with no significant improvement. This indicates that the model failed to capture any meaningful inverse mapping structure. The results highlight that the MLP struggled to achieve meaningful improvements, with persistent high loss across both training and validation, further reinforcing the complexity of the one-to-many prediction problem.

Baseline models were evaluated by comparing predicted configurations to the corresponding ground truth inputs in the dataset. While this does not account for the one-to-many nature of the problem, it provides a standardized measure of model accuracy. We acknowledge that a more meaningful comparison would involve verifying whether the predicted inputs yield the correct output, which is how our Teacher-Student and reinforcement learning (RL) models are evaluated.

\subsection{Teacher-Student Framework Performance}

Building on the limitations observed in baseline inverse modeling, we next implemented the forward Teacher model, which serves as the foundation of our Teacher-Student framework. The Teacher is trained to learn the many-to-one mapping from the 13 input features—tempering time, temperature, and chemical composition—to the target variable: final hardness (HRC).

We used a Multi-Layer Perceptron (MLP) architecture for the Teacher model. The training process was carried out using standard supervised learning, with mean squared error (MSE) as the loss function.

\begin{figure}
  \centering
  \begin{subfigure}[t]{0.48\linewidth}
    \centering
    \includegraphics[width=\linewidth]{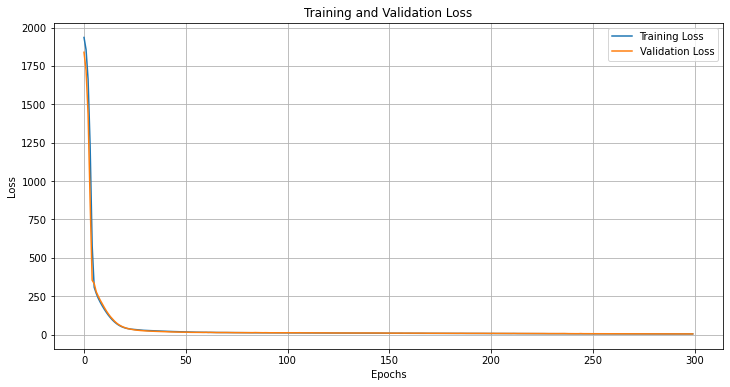}  
    \caption*{(a)}
  \end{subfigure}
  \hfill
  \begin{subfigure}[t]{0.48\linewidth}
    \centering
    \includegraphics[width=\linewidth]{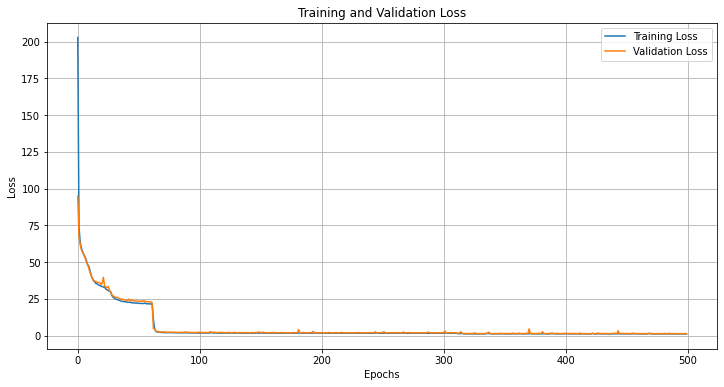}  
    \caption*{(a)}
  \end{subfigure}
  \caption{Training and validation loss curves. (a) The forward (Teacher) MLP model converges quickly within the first 50 epochs, with both losses approaching zero, demonstrating strong generalization and suitability as a reference for inverse training. (b) The inverse (Student) model effectively learns the mapping over 500 epochs, with both training and validation losses sharply decreasing and stabilizing.}

  \label{fig:teacher_student}
\end{figure}

As shown in Figure~\ref{fig:teacher_student}a, the model converged rapidly and effectively. Both training and validation loss decreased sharply within the first few epochs and stabilized at low values, demonstrating excellent generalization and predictive performance. This strong forward model is critical, as it serves as a consistent, differentiable reference for training the inverse Student model in the next phase.

Once the Teacher model was trained and fixed, we proceeded to train the Student model to solve the inverse problem: predicting process parameters from a target hardness value. The Student model was trained in a collaborative loop with the Teacher, allowing it to learn valid inverse mappings despite the many-to-one nature of the problem. Each training iteration proceeds as follows:

\begin{itemize}
  \item A batch of random target hardness values is sampled.
  \item The Student predicts the corresponding 13 input features, including tempering parameters and elemental composition.
  \item These predicted inputs are passed through the Teacher model, which outputs predicted hardness values.
  \item A loss is calculated by comparing the Teacher’s output to the original target.
  \item The loss is backpropagated to update the Student model's weights.
\end{itemize}

This iterative feedback mechanism continues until convergence. As shown in Figure~\ref{fig:teacher_student}b, the training and validation loss both decrease rapidly and stabilize at low values, indicating that the Student successfully learns to produce inputs that yield accurate hardness predictions when passed through the Teacher.

\begin{figure}[t]
  \centering
  \includegraphics[width=0.65\linewidth]{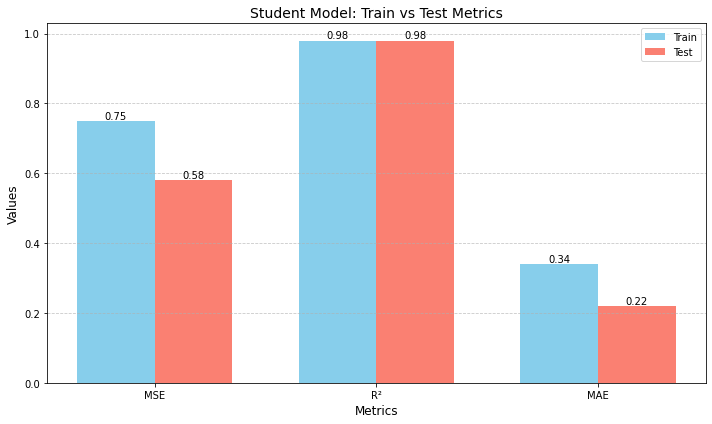}
  \caption{Evaluation metrics for the Student model on training and test data. The model achieved high $R^2$ values (0.98) and low mean squared and absolute errors, demonstrating accurate and generalizable inverse predictions when supervised by the Teacher model.}
  \label{fig:evaluation}
\end{figure}

Evaluation metrics, shown in Figure~\ref{fig:evaluation}, further confirm the Student’s effectiveness. The model achieved high $R^2$ scores of 0.98 on both training and test sets, with relatively low mean squared error (MSE) and mean absolute error (MAE) values. Importantly, the predicted input values do not need to match dataset entries exactly—due to the many-to-one nature of the mapping—but only need to generate the correct target output. This flexibility is a key strength of the Teacher-Student framework. It allows the Student model to explore the solution space beyond the training data, while still producing consistent, functionally accurate results.

In contrast to the approaches described previously—such as Bundle Networks \cite{courts2021} or mustGAN \cite{yurt2021} our method is designed to be both simple and data-efficient. Rather than employing complex generative models, recurrent structures, or transformers, we adopt a purely feedforward architecture based on MLPs with residual connections and ELU activations. Unlike previous methods that require large datasets and intricate training regimes, our Teacher-Student framework can learn effective inverse mappings from limited data. Additionally, our method avoids probabilistic modeling, employing a deterministic Student model supervised by a fixed Teacher MLP. This architectural simplicity, combined with a training strategy focused on consistency rather than reconstruction, allows our model to produce multiple functionally valid input configurations without resolving all ambiguity. These distinctions make our approach especially well-suited for industrial applications.

\subsection{Comparison with Reinforcement Learning Approach}

To benchmark the performance of our Teacher-Student approach, we implemented a model-free reinforcement learning (RL) agent using the TD3 (Twin Delayed Deep Deterministic Policy Gradient) algorithm. Like the Student model, the RL agent was tasked with solving the inverse problem: predicting 13 process parameters that would yield a specified target hardness value.

The training setup used the trained Teacher model as the environment. At each timestep:

\begin{itemize}
  \item The agent receives a target hardness value.
  \item It predicts a set of 13 input variables (time, temperature, composition).
  \item These inputs are passed to the Teacher, which returns a predicted hardness.
  \item A reward is computed as the negative mean squared error (MSE) between the predicted and actual hardness.
  \item This reward is used to update the agent’s policy via TD3.
\end{itemize}
\FloatBarrier  
\begin{table*}[htbp]
  \caption{Performance comparison of inverse modeling approaches. The Teacher-Student framework achieves the lowest MSE and fastest convergence, outperforming both traditional models (Random Forest, baseline MLP) and the RL-based TD3 agent in accuracy and efficiency.}
  \label{tab:model-comparison}
  \centering
  \begin{tabular}{lccc}
    \toprule
    \textbf{Model} & \textbf{MSE} & \textbf{Training Time} & \textbf{Notes} \\
    \midrule
    Teacher-Student   & 0.52     & 2.3 min      & Best overall performance \\
    RL (TD3)          & 4.84     & 27.7 min     & Slower training, higher error \\
    Random Forest     & 620.54   & $\sim$1.2 min & Poor generalization \\
    MLP (baseline)    & 15730    & $\sim$3.5 min & Failed to converge \\
    \bottomrule
  \end{tabular}
\end{table*}

\begin{figure}[t!]
  \centering
  \begin{subfigure}[t]{0.48\linewidth}
    \centering
    \includegraphics[width=\linewidth]{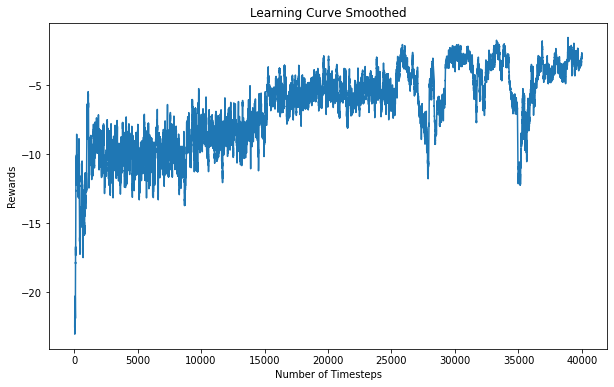}
    \caption*{(a)}
  \end{subfigure}
  \hfill
  \begin{subfigure}[t]{0.48\linewidth}
    \centering
    \includegraphics[width=\linewidth]{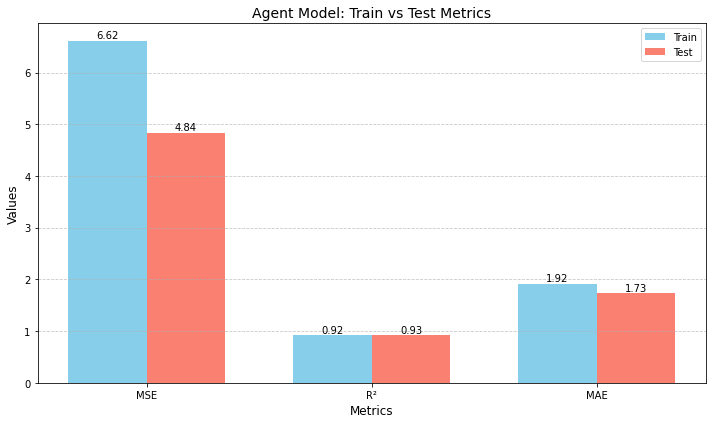}
    \caption*{(b)}
  \end{subfigure}
  \caption{Performance of the reinforcement learning (RL) approach. (a) Smoothed learning curve of the TD3 agent over 40,000 timesteps, showing gradual policy improvement with typical fluctuations. (b) Evaluation metrics indicate that while the RL model achieves $R^2$ scores above 0.9, its MSE and MAE are significantly higher than those of the Teacher-Student model, highlighting its lower accuracy and efficiency.}
  \label{fig:rl}
\end{figure}

The RL agent was trained for 40{,}000 steps, compared to 15{,}000 steps for the Student model. As shown in Figure~\ref{fig:rl}a, the smoothed reward curve demonstrates gradual performance improvement, indicating convergence. However, the learning process is noticeably more volatile and prolonged.

In terms of quantitative performance, Figure~\ref{fig:rl}b shows that the RL agent achieved reasonable accuracy, with $R^2$ values around 0.92–0.93, and moderate error rates (MSE = 4.84, MAE = 1.73 on the test set). Despite these results, the RL agent was clearly outperformed by the Student model, which achieved significantly better accuracy with far less computational cost.

Notably, training the RL agent took 27.7 minutes, while the Teacher-Student system required only 2.3 minutes. These findings highlight the efficiency, stability, and predictive quality of our proposed framework over traditional reinforcement learning approaches for inverse process modeling.

\section{Conclusions and Future Work}

In this work, we tackled the inverse prediction problem in steel heat treatment, where multiple input configurations can lead to the same final hardness, creating a non-invertible, many-to-one mapping that poses a major challenge for conventional regression methods. To address this, we introduced a Teacher-Student learning framework in which a forward model (Teacher) predicts hardness from process parameters, and an inverse model (Student) learns to infer valid input configurations from target hardness values through iterative supervision.

Our approach was evaluated on a real-world dataset of tempered steel and demonstrated superior performance compared to both standard regression techniques and a reinforcement learning baseline, achieving higher accuracy and significantly lower computational cost. However, while our Student model is capable of generating functionally correct inverse configurations, we did not explicitly enforce or measure diversity in the predicted inputs. Because the loss is computed on the Teacher’s output, the Student is not constrained to match any specific input sample from the dataset. In principle, this allows it to converge on multiple valid solutions per target value. However, we acknowledge that without explicit diversity-promoting mechanisms (e.g., entropy regularization, sampling strategies or loss enforcing), the model may still collapse toward a subset of the input space. Quantifying the variance of predicted configurations is an important direction for future work, especially in scenarios requiring solution diversity or process flexibility.

Looking forward, several directions can extend this work. Incorporating uncertainty modeling could allow the Student to express confidence in its predictions. Attention mechanisms or transformer-based architectures may improve both accuracy and interpretability. Applying the framework to other material properties or processes would test its generalization ability, while integration into real-time control systems could enable dynamic process optimization. Finally, combining this data-driven approach with physics-informed constraints offers a path toward more robust and explainable models for complex inverse problems in materials science.

\subsection*{Acknowledgments}
The authors wish to thank Thomas Albrecht for his support, fruitful discussions, and useful advice.

\subsection*{Disclosure of Interests}
The first author is pursuing a PhD at Deggendorf Institute of Technology in collaboration with Krones AG, which funded the research presented in this article. The second author, a faculty member at Deggendorf Institute of Technology, contributed in an academic supervisory capacity. The authors declare that they have no other competing interests.

\section*{Declaration on Generative AI}
 During the preparation of this work, the author(s) used ChatGPT (GPT-4) in order to: assist with language editing, LaTeX formatting, and figure caption refinement. After using this tool, the author(s) reviewed and edited the content as needed and take(s) full responsibility for the publication’s content.

\bibliography{sample-ceur}

\end{document}